\documentclass[a4paper]{article}

\usepackage{INTERSPEECH2018}
\usepackage[utf8]{inputenc}

\usepackage{amsmath,graphicx}
\usepackage{tikz,booktabs,float,color,adjustbox}
\usetikzlibrary{arrows}
\usepackage[utf8]{inputenc}
\usepackage{pgfplots}
\pgfplotsset{width=10cm,compat=1.9}

\definecolor{kit-blue100}{cmyk}{.8,.5.,0,0}
\definecolor{kit-blue70}{cmyk}{.56,.35,0,0}
\definecolor{kit-blue50}{cmyk}{.4,.25,0,0}
\definecolor{kit-blue30}{cmyk}{.24,.15,0,0}
\definecolor{kit-blue15}{cmyk}{.12,.075,0,0}

\definecolor{kit-green100}{cmyk}{1,0,.6,0}
\definecolor{kit-green70}{cmyk}{.7,0,.42,0}
\definecolor{kit-green50}{cmyk}{.5,0,.3,0}
\definecolor{kit-green30}{cmyk}{.3,0,.18,0}
\definecolor{kit-green15}{cmyk}{.15,0,.09,0}

\definecolor{kit-merge70}{cmyk}{.63,.175,.21,0}

\definecolor{cmu-red100}{cmyk}{0,1.0,.79,.2}
\definecolor{cmu-red15}{cmyk}{0,.15,.1185,.03}

\title{Neural Language Codes for Multilingual Acoustic Models}
\name{Markus Müller$^1$, Sebastian Stüker$^1$, and Alex Waibel$^{1,2}$ \thanks{This work was realized in the framework of the ANR-DFG project BULB (ANR-14-CE35-002).}}

\address{
  $^1$Karlsruhe Institute of Technology, Karlsruhe, Germany\\
  $^2$Carnegie Mellon University, Pittsburgh PA, USA}
\email{\{m.mueller,sebastian.stueker,alexander.waibel\}@kit.edu}

\begin{document}
\maketitle
\begin{abstract}
Multilingual Speech Recognition is one of the most costly AI problems, because each language (7,000+) and even different accents require their own acoustic models to obtain best recognition performance.  Even though they all use the same phoneme symbols, each language and accent imposes its own coloring or “twang”.  Many adaptive approaches have been proposed, but they require further training, additional data and generally are inferior to monolingually trained models.  In this paper, we propose a different approach that uses a large multilingual model that is \emph{modulated} by the codes generated by an ancillary network that learns to code useful differences between the “twangs” or human language. 

We use Meta-Pi networks \cite{hampshire1990meta,hampshire1992meta} to have one network (the language code net) gate the activity of neurons in another (the acoustic model nets).  Our results show that during recognition multilingual Meta-Pi networks quickly adapt to the proper language coloring without retraining or new data, and perform better than monolingually trained networks.
The model was evaluated by training acoustic modeling nets and modulating language code nets jointly and optimize them for best recognition performance.   
\end{abstract}
\section{Introduction}
Multilingual speech recognition is a challenging problem, because each of the 7,000 living languages and even dialects or accents require own acoustic model for optimal performance.
In low-resource scenarios when only a limited amount of training data from the target language is available, the use of data from additional languages is a well established method to improve the system performance.
Training on more data from multiple languages allows the models to generalize better.
If enough training data is available, monolingual systems display the best performance and systems trained on a combination of languages may even have an inferior performance.
Training on multiple languages introduces additional ambiguity.
Corpora from different languages may also be annotated in different manners, e.g. using different phoneme sets.
Language adaptation methods are required to improve the performance of such acoustic models.

Language adaptation poses different problems than speaker adaptation.
Collecting data from several hundreds of speakers is feasible, while collecting data from the same amount of languages is next to impossible.
Training on a small number of languages does not enable the network to generalize across languages.
Similar issues for speaker adaptation were reported in the early days of automatic speech recognition (ASR) systems, as these systems were trained on data from an equally low number of speakers.

In this work, we will show how acoustic models can be adapted to languages in a multilingual setting.
We have proposed adaptation methods based on language features  \cite{mueller2018icassp,mueller2016} and now propose a new approach that extends our previous work.
We will incorporate two novel aspects: a) the use of adaptive neural language codes (NLCs), which are based on language feature vectors (LFVs) \cite{mueller2016}, but can be adapted during acoustic model training and b) a network superstructure based on Meta-PI \cite{hampshire1990meta,hampshire1992meta}, which allows to use pre-trained monolingual subnets in a multilingual system.
By combining both approaches, we are able to train a multilingual acoustic models which do not only achieve monolingual performance, also improve beyond monolingual systems.

This paper is organized as follows: In Section \ref{sec:relwork} we provide an overview of related work in the field.
The extraction of neural language codes (NLCs) is outlined in Section \ref{sec:nlc} and the main network architecture in Section \ref{sec:arch}.
We describe our experimental setup in Section \ref{sec:expsetup} and the results in Section \ref{sec:results}.
This paper concludes with Section \ref{sec:conclusion}, where we also provide an outlook to future work.
\section{Related Work}
\label{sec:relwork}
\subsection{Multi- and Crosslingual Speech Recognition Systems}
\label{sec:relwork:subsec:gmm}
Prior to the emergence of neural networks, ASR systems were typically built using a GMM/HMM based approach.
Methods for training/adapting such systems cross- and multilingually were proposed to handle data sparsity \cite{schultz1997fast,schultz1998multilingual,stuker2009acoustic}.
The process of clustering context-independent phones into context-dependent ones can also be adapted to account for cross- and multilinguality \cite{ip_stueker2008a}.
\subsection{Neural Network Adaptation}
Supplying additional features to neural networks for adaptation to certain conditions is a common technique.
A very common method for speaker adaptation is using i-Vectors \cite{dehak2011front,saon2013speaker} which are a low-dimensional representation of speaker and/or channel characteristics.
Based on these vectors, speaker adaptive networks can be trained \cite{miao2014towards}.
These low-dimensional representations can also be extracted by a neural network and are called Bottleneck Speaker Vectors (BSVs) \cite{huang2015investigation}.

Based on this idea, we used a similar method for adapting DNNs to multiple languages.
First, we utilized only the language identity, encoded via one-hot encoding \cite{mueller2015}.
This approach was refined by using Language Feature Vectors (LFVs) \cite{mueller2016}.
Like BSVs, LFVs were extracted using a neural network.
In comparison to using the language identity alone, LFVs enable better language adaptation which results in lower WERs.
\subsection{RNN Based ASR Systems}
Artificial neural network (ANN) based ASR systems have gained a lot of research interest in recent years.
Due to increased computing capabilities, more complex architectures can be trained on even more data.
A novel approach for building systems is to use recurrent neural networks (RNNs) trained using the connectionist temporal classification (CTC) loss function \cite{graves2006connectionist}.
Being a powerful tool for sequence classification, RNNs are able to model temporal contexts implicitly.
No explicit modelling of context-dependent targets as in traditional systems is required.
Phones, graphemes or both can be used as acoustic modeling units \cite{chen2014joint}.
Training on whole words is also possible, given enough training data \cite{soltau2016neural}.
Other approaches to train a monolithic acoustic model on multiple languages were also proposed \cite{watanabe2017language}.
\section{Neural Language Codes}
\label{sec:nlc}
We proposed the use of features encoding language properties \cite{mueller2016}.
These so-called language feature vectors (LFVs) enable language adaptation of neural networks.
The network architecture is shown in Figure \ref{fig:lfvnet}.
To extract those features, we trained a deep neural network (DNN) for language identification.
This network featured a bottleneck as second-to-last layer.
After training, the layers after this layer were discarded and the output activations of the bottleneck layer were taken as LFVs.
\tikzstyle{layer}=[draw=black,fill=kit-green70]
\tikzstyle{layerinout}=[draw=black,fill=black!30]
\tikzstyle{layerbnf}=[draw=black,fill=black!30]
\tikzstyle{layerlfv}=[draw=black,fill=kit-blue70]
\tikzstyle{dots}=[draw=black,fill=black]
\tikzstyle{freezebg}=[draw=cmu-red100!50,very thick,fill=cmu-red100!20]
\tikzstyle{dots}=[draw=black,fill=black]
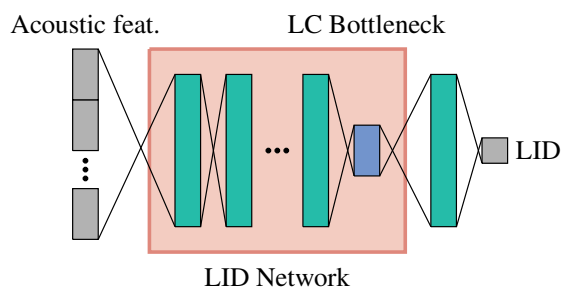
\begin{figure}[htbp]
  \centering
  \resizebox{0.45\textwidth}{!}{
  \begin{tikzpicture}[scale=0.6]

  \draw (4.25, 1) node {Acoustic feat.};

	\fill[freezebg] (5.5,0.5) -- (10.5,0.5) -- (10.5,-3.5) -- (5.5,-3.5) -- (5.5,0.5);
	\draw (8,-4) node {LID Network};

  \fill[layerbnf] (4,-1.5) coordinate(l5_1bl) -- (4.5,-1.5) coordinate(l5_1br) -- (4.5,-0.5) coordinate(l5_1tr) -- (4,-0.5) coordinate(l5_1tl) -- (4,-1.5);

  \draw[dots] (4.25,-1.7) circle (0.045);
  \draw[dots] (4.25,-1.875) circle (0.045);
  \draw[dots] (4.25,-2.05) circle (0.045);

  \fill[layerbnf] (4,-3.25) coordinate(l5_2bl) -- (4.5,-3.25) coordinate(l5_2br) -- (4.5,-2.25) coordinate(l5_2tr) -- (4,-2.25) coordinate(l5_2tl) -- (4,-3.25);
  \fill[layerbnf] (4,-0.5) coordinate(l5bl) -- (4.5,-0.5) coordinate(l5br) -- (4.5,0.5) coordinate(l5tr) -- (4,0.5) coordinate(l5tl) -- (4,-0.5);

  \fill[layer] (6,-3) coordinate(l6bl) -- (6.5,-3) coordinate(l6br) -- (6.5,0) coordinate(l6tr) -- (6,0) coordinate(l6tl) -- (6,-3);
  \fill[layer] (7,-3) coordinate(l7bl) -- (7.5,-3) coordinate(l7br) -- (7.5,0) coordinate(l7tr) -- (7,0) coordinate(l7tl) -- (7,-3);

  \draw[dots] (8,-1.5) circle (0.045);
  \draw[dots] (8.175,-1.5) circle (0.045);
  \draw[dots] (7.825,-1.5) circle (0.045);

  \fill[layer] (8.5,-3) coordinate(l11bl) -- (9,-3) coordinate(l11br) -- (9,0) coordinate(l11tr) -- (8.5,0) coordinate(l11tl) -- (8.5,-3);
  \fill[layerlfv] (9.5,-2) coordinate(l12bl) -- (10,-2) coordinate(l12br) -- (10,-1) coordinate(l12tr) -- (9.5,-1) coordinate(l12tl) -- (9.5,-2);

  \draw (9.75, 1) node {LC Bottleneck};

  \fill[layer] (11,-3) coordinate(l13bl) -- (11.5,-3) coordinate(l13br) -- (11.5,0) coordinate(l13tr) -- (11,0) coordinate(l13tl) -- (11,-3);

  \fill[layerinout] (12,-1.75) coordinate(l14bl) -- (12.5,-1.75) coordinate(l14br) -- (12.5,-1.25) coordinate(l14tr) -- (12,-1.25) coordinate(l14tl) -- (12,-1.75);

  \draw (13.1,-1.5) node {LID};

  \draw (l5tr) -- (l6bl);
  \draw (l5_2br) -- (l6tl);

  \draw (l6tr) -- (l7bl);
  \draw (l6br) -- (l7tl);
  \draw (l11tr) -- (l12bl);
  \draw (l11br) -- (l12tl);
  \draw (l12tr) -- (l13bl);
  \draw (l12br) -- (l13tl);
  \draw (l13tr) -- (l14bl);
  \draw (l13br) -- (l14tl);

  \end{tikzpicture}
  }
  \caption{Language Feature Vectors (LFVs) network architecture}
    \label{fig:lfvnet}
\end{figure}

Adaptation to languages is not as signal related as speaker adaptation.
Properties like the length of the vocal tract or the fundamental frequency manifest themselves in the spectrum. Thus, speaker features are typically added at the acoustic feature level.
For language adaptation, we showed that adding LFVs deeper into the network \cite{mueller2018icassp} does improve the performance, and also optimized this approach \cite{essv2018}.
Key is a method called modulation, which is based on Meta-PI networks \cite{hampshire1990meta,hampshire1992meta}.
These networks feature Meta-PI connections which allow to modulate the output of a neural unit by multiplication with a coefficient.
Based on this coefficient, the unit's output value will either be increased or decreased.

As such, in order to modulate the outputs of a whole layer, the number of coefficients has to match the number of outputs.
In our experiments we therefore chose the number of units per layer to be a multiple of the dimensionality of LFVs.
By stacking LFVs, we could then match the dimensionality of the LFVs to the dimensionality of the layer's outputs.
But this also means that the outputs of multiple neurons will be modulated with the same coefficient or the layer's outputs will be divided into multiple groups.
The outputs of each unit will be emphasized or attenuated based on the modulation with language codes.
The units will in turn become sensitive to language properties.
Modulating the outputs in this manner can be considered ``intelligent dropout'', as the connections between units are altered in a systematic way, in comparison to randomly dropping entire connections during dropout training \cite{srivastava2014dropout}.

The division of the layer's outputs into groups of equal size for modulation is arbitrary.
A better solution would be to break free of this fixed grouping imposed by stacking of the same features multiple times and to furthermore allow updates to the modulation coefficients.
We trained a neural language codes (NLC) network to mitigate these issues.
The network's architecture is shown in Figure \ref{fig:nlcnet}.
It is trained to generate the stacked LFV output, using both LFVs and acoustic features (multilingual bottleneck features, ML-BNFs) as input.
During the training process, it will most likely learn to simply forward the LFVs to the appropriate outputs and to ignore the acoustic features.
This network will later be integrated into the network superstructure and adapted based on the new task.
By this adaptation, it will learn to transform language feature vectors encoding language properties into neural language codes, which still encode language characteristics, but are now optimized towards speech recognition.
\tikzstyle{layerainput}=[draw=black,very thick,fill=kit-green70]
\tikzstyle{layerlinput}=[draw=black,very thick,fill=kit-blue70]
\tikzstyle{layerminput}=[draw=black,very thick,fill=kit-merge70]
\tikzstyle{layeroutput}=[draw=black,very thick,fill=black!10]
\tikzstyle{dots}=[draw=black,fill=black]
\tikzstyle{myarrows}=[black,line width=1mm,fill=white,preaction={-triangle 90,thin,draw,shorten >=-1mm}]
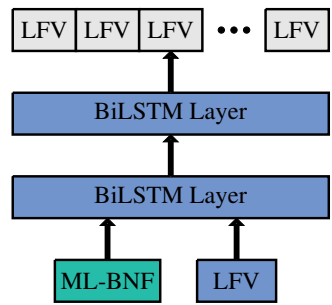
\begin{figure}[!h]
\centering
\resizebox{0.25\textwidth}{!}{%
\begin{tikzpicture}[scale=0.36]

  \fill[layerainput] (-2.5,1) -- (2.5,1) -- (2.5,-1) -- (-2.5,-1) -- (-2.5,1);
  \node[align=center,font=\large,rotate=0] at (0,0) {ML-BNF};

  \fill[layerlinput] (4.25,1) -- (8,1) -- (8,-1) -- (4.25,-1) -- (4.25,1);
  \node[align=center,font=\large,rotate=0] at (6.125,0) {LFV};

  \fill[layerlinput] (-4.5,5) -- (10.5,5) -- (10.5,3) -- (-4.5,3) -- (-4.5,5);
  \node[align=center,font=\large,rotate=0] at (3,4) {BiLSTM Layer};

  \fill[layerlinput] (-4.5,9) -- (10.5,9) -- (10.5,7) -- (-4.5,7) -- (-4.5,9);
  \node[align=center,font=\large,rotate=0] at (3,8) {BiLSTM Layer};

  \fill[layeroutput] (-4.5,13) -- (-1.5,13) -- (-1.5,11) -- (-4.5,11) -- (-4.5,13);
  \node[align=center,font=\large,rotate=0] at (-3,12) {LFV};

  \fill[layeroutput] (-1.5,13) -- (1.5,13) -- (1.5,11) -- (-1.5,11) -- (-1.5,13);
  \node[align=center,font=\large,rotate=0] at (0,12) {LFV};

  \fill[layeroutput] (1.5,13) -- (4.5,13) -- (4.5,11) -- (1.5,11) -- (1.5,13);
  \node[align=center,font=\large,rotate=0] at (3,12) {LFV};

  \draw[dots] (5.4,12) circle (0.15);
  \draw[dots] (6.0,12) circle (0.15);
  \draw[dots] (6.6,12) circle (0.15);

  \fill[layeroutput] (7.5,13) -- (10.5,13) -- (10.5,11) -- (7.5,11) -- (7.5,13);
  \node[align=center,font=\large,rotate=0] at (9,12) {LFV};

  \draw[myarrows] (0,1) -- (0,2.75);
  \draw[myarrows] (6.125,1) -- (6.125,2.75);

  \draw[myarrows] (3,5) -- (3,6.75);
  \draw[myarrows] (3,9) -- (3,10.75);

\end{tikzpicture}
}
\caption{Neural Language Codes (NLC) network architecture, pre-trained to stack LFVs}
\label{fig:nlcnet}
\end{figure}
\section{Network Architecture}
\label{sec:arch}
%
%
In order to leverage the full potential of monolingual models in our multilingual setup, we explored how monolingual models can be integrated into our network architecture.
One method is to use a network architecture based on Meta-PI \cite{hampshire1990meta,hampshire1992meta}.
The authors presented an approach where parts of the network were trained on different aspects of the same problem.
These subnets would then be combined to a larger network superstructure with a trainable component to determine the mixture weights of the outputs from the individual networks.

Based on this setup, we opted for a similar network architecture, where language dependent subnets were pre-trained and then combined to a larger network.
On top of these subnets, we would use our default network architecture: A two part BiLSTM network, with the output of the first block being modulated by language features.
%
%
\definecolor{mypink2}{RGB}{219, 48, 122}

\tikzstyle{layerainput}=[draw=black,very thick,fill=kit-green70]
\tikzstyle{layerlinput}=[draw=black,very thick,fill=kit-blue70]
\tikzstyle{layerminput}=[draw=black,very thick,fill=kit-merge70]
\tikzstyle{layeroutput}=[draw=black,very thick,fill=black!10]
\tikzstyle{dots}=[draw=black]

\tikzstyle{myarrows}=[black,line width=1mm,fill=white,preaction={-triangle 90,thin,draw,shorten >=-1mm}]

\begin{figure}[!h]
\centering
\resizebox{0.35\textwidth}{!}{%
\begin{tikzpicture}[scale=0.36]

\fill[layerainput] (-2.5,1) -- (2.5,1) -- (2.5,-1) -- (-2.5,-1) -- (-2.5,1);
\node[align=center,font=\large,rotate=0] at (0,0) {ML BNF};

\fill[layerlinput] (4,1) -- (8,1) -- (8,-1) -- (4,-1) -- (4,1);
\node[align=center,font=\large,rotate=0] at (6,0) {LFV};

\draw[myarrows] (0,1.1) -- (-7.5,3.8);
\draw[myarrows] (0,1.1) -- (-3.5,3.8);
\draw[myarrows] (0,1.1) -- (0.5,3.8);
\draw[myarrows] (0,1.1) -- (4.5,3.8);
\draw[myarrows] (0,1.1) -- (9,4.8);

\draw[myarrows] (6.0,1) -- (9.2,4.8);

\fill[layerainput] (-9,8) -- (-6,8) -- (-6,4) -- (-9,4) -- (-9,8);
\node[align=center,font=\small\bf,rotate=0] at (-7.5,6) {DE\\Phone};

\fill[layerainput] (-5,8) -- (-2,8) -- (-2,4) -- (-5,4) -- (-5,8);
\node[align=center,font=\small\bf,rotate=0] at (-3.5,6) {FR\\Phone};

\fill[layerainput] (-1,8) -- (2,8) -- (2,4) -- (-1,4) -- (-1,8);
\node[align=center,font=\small\bf,rotate=0] at (0.5,6) {TR\\Phone};

\fill[layerainput] (3,8) -- (6,8) -- (6,4) -- (3,4) -- (3,8);
\node[align=center,font=\small\bf,rotate=0] at (4.5,6) {EN\\Char};

\draw[myarrows] (9,14.63) -- (9,6);

\fill[layerlinput] (7,7) -- (11,7) -- (11,5) -- (7,5) -- (7,7);
\node[align=center,font=\small\bf,rotate=0] at (9,6) {NLC};

\draw[myarrows] (9,14.5) -- (-0.5,14.5);

\draw[myarrows] (-7.5,8) -- (-7.5,9.7);
\draw[myarrows] (-3.5,8) -- (-3.5,9.7);
\draw[myarrows] (0.5,8) -- (0.5,9.7);
\draw[myarrows] (4.5,8) -- (4.5,9.7);

\fill[layerainput] (-9,13) -- (6,13) -- (6,10) -- (-9,10) -- (-9,13);
\node[align=center,font=\large,rotate=0] at (-1.5,11.5) {BiLSTM Part 1};

\draw[myarrows] (-1.5,13.1) -- (-1.5,13.5);
\draw[dots,line width=0.5mm] (-1.5,14.5) circle (0.75);
\node[align=center,font=\large\bf,rotate=0] at (-1.47,14.35) {*};
\draw[myarrows] (-1.5,15.2) -- (-1.5,15.8);

\fill[layerminput] (-9,19) -- (6,19) -- (6,16) -- (-9,16) -- (-9,19);
\node[align=center,font=\large,rotate=0] at (-1.5,17.5) {BiLSTM Part 2};

\draw[myarrows] (-1.5,19.0) -- (-1.5,20.7);

\fill[layeroutput] (-7,23) -- (4,23) -- (4,21) -- (-7,21) -- (-7,23);
\node[align=center,font=\large,rotate=0] at (-1.5,22) {Output Layer};

\end{tikzpicture}
}
\caption{Network superstructure based on Meta-PI, using Neural Language Codes (NLC) for network modulation.}
\label{fig:metapi}
\end{figure}
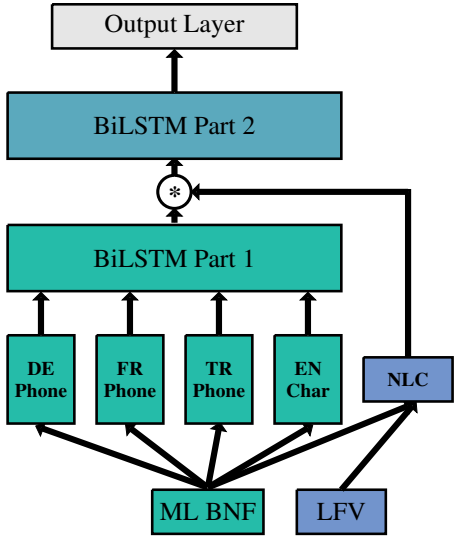
\subsection{Monolingual Sub Networks}
Each monolingual network was trained on a single language only.
We trained two networks per language: One using graphemic and one using phonemic targets.
We evaluated using networks having both only a quarter or half the amount of LSTM cells per layer in comparison to the main network.
We also reduced the number of layers from 4 to 3.
This network size was selected to limit the number of parameters to not only prevent over-fitting, but also due to memory constraints of the GPUs used for training.
\subsection{Main Network}
Our main network featured 4 BiLSTM layers with 420 cells each.
It was divided into two parts of 2 layers each, with the modulation being applied between these parts.
For applying the modulation, we first combined to output of the cells for each direction pairwise by taking the maximum \cite{essv2018}.
The combined outputs were then modulated with NLCs.
\subsection{Network Training}
Initially, the parameters of the pre-trained subnets were loaded.
This includes all the language specific nets, as well as the NLC network.
The weights of the main network were randomly initialized.
The entire network was then jointly trained on the combined graphemic targets of the 4 languages.
This joint training allows the individual subnets to adapt to the global task.
\section{Experimental Setup}
\label{sec:expsetup}
We used data from the Euronews corpus \cite{gretter2014euronews}, which consists of TV broadcast news recordings from 10 languages.
While we used data from 4 languages (English, French, German, Turkish) to train our ASR system, different subsets were used for training the subnets, as indicated in each subsection.
The data was filtered removing utterances shorter than 1s or a transcript with more than 639 symbols\footnote{Internal limitation within the implementation of CUDA/warp-ctc, see: https://github.com/baidu-research/warp-ctc, accessed 2018-03-16}.
This corpus features only very basic annotation of noises with a single noise marker representing any type of noise, e.g. music, human and non-human noises or unintelligible speech.
Utterances containing only noise were therefore discarded.
In total, 50h of data per language remained after the cleaning process, and this dataset was split into 45h of training and 5h of test data.

As acoustic features, we used multilingual bottleneck features (ML-BNFs).
They were trained as part of previous experiments \cite{mueller2016} on Euronews data, using a combination of 5 languages (French, German, Italian, Russian, Turkish).
The network is fed a combination of logMel and tonal features (FFV \cite{kornel:ffv}, pitch \cite{kjell:da}), extracted via a 32ms window with a 10ms frame-shift.
\subsection{Acoustic Units}
Phones and graphemes were used as acoustic modelling units.
The pronunciation dictionaries were created automatically using MaryTTS \cite{schroder2003german}.
We mapped the phones of each language using the articulatory features embedded in MaryTTS' language definition files to create a global phone set.
In addition, we used a token representing word boundaries.
\subsection{Language Dependent Subnets}
We trained monolingual networks for each language, using both graphemic and phonetic targets.
Each network featured 3 bi-directional layers with either 105 or 210 LSTM cells.
The number of cells was chosen based on the size of the main network and we opted for using layers with half or a quarter of LSTM cells.
As input features, we used ML-BNFs.
A feed-forward layer was used as output layer to map the outputs of the last hidden layer to the output targets.
The networks were trained using the CTC loss function, stochastic gradient descent (SGD) and Nesterov momentum \cite{sutskever2013importance} with a factor of $0.9$.
The utterances were sorted ascending by length to stabilize the training, as shorter utterances are easier to align.
After training, the output layers were discarded and the outputs of the last hidden layer were used in our network superstructure.
By using the features extracted by the last hidden layer instead of the classification result, this setups becomes agnostic towards the set of acoustic units each network uses.
As we were using bi-directional LSTM layers, each cell outputs two coefficients, one for each direction.
Based on previously reported experiments \cite{essv2018}, we opted for taking the pairwise maximum value for each direction, in the notion of maxpool / maxout \cite{goodfellow2013maxout} layers.
\subsection{Neural Language Codes}
For modulating our main network, we extract NLCs based on a two layer, bi-directional LSTM network with 420 LSTM cells per layer.
As input features, we supplied both ML-BNFs and LFVs.
The network was trained to output stacked LFVs using mean squared error as loss function.
Given the bi-directional nature of the network, we chose to sum the values pairwise for each direction.
Applying the sum over the maximum is potentially more stable in approximating real valued outputs.
Preliminary experiments showed that no big differences between both methods exist.
\subsection{RNN/CTC Network}
The main network consisted of 4 layers with 420 BiLSTM cells each.
The number of cells was chosen to be a multiple of the dimensionality of the LFVs.
The network was split into two parts with 2 layers each.
The first part used the combined outputs of the language dependent subnets as input features.
The outputs were merged pairwise using the same maxpool strategy as the subnets.
The modulation is then applied prior to feeding the outputs into the second part.
A feed-forward layer then mapped the outputs of the last LSTM layer to the targets.
\subsection{Grapheme Based RNN LM}
For decoding, we used a character based RNN LM.
It was trained as described in \cite{zenkel2017comparison} on the transcripts of the training utterances (110k sentences).
We first trained a baseline LM with a single layer of 1024 BiLSTM units.
We later re-fined it (``new LM'' in Table \ref{tab:lmdecode}) by optimizing the number of BiLSTM cells.
In a series of experiments we determined 512 BiLSTM cells to be the optimal number, resulting in the lowest WER.
\subsection{Training Strategy}
The network superstructure was trained in multiple steps.
First, the language dependent subnets were trained individually, so was the NLC network.
These networks were then combined with the main network.
We evaluated using different combinations of source nets, mixing nets trained on graphemic and phonemic targets (as shown in Figure \ref{fig:metapi}).
The entire architecture was trained jointly, allowing updates to the parameters of all networks.
Similar to training the subnets, SGD and Nesterov momentum \cite{sutskever2013importance} with a factor of $0.9$ were used for training.
But given the increased parameter count, we also applied Dropout training with a factor of $0.2$ to prevent over fitting.
\subsection{Evaluation}
The network performance was evaluated using both character error rate (CER) and word error rate (WER).
For decoding, we used the same procedure as in \cite{graves2006connectionist} and greedily searched for the best path.
WER results were obtained by performing a decoding using the RNN LM.
We compared training multilingual networks for various conditions.
As contrasting experiment, we trained an English monolingual system.
\section{Results}
\label{sec:results}
We first trained monolingual subnets on graphemes and phonemes using different layer sizes and evaluated their performance.
Next, we combined these language dependent subnets in our network superstructure.
\subsection{Monolingual Subnets}
The CERs of our individual monolingual subnets are shown in Table \ref{tab:subnets}.
Training networks with only 105 BiLSTM cells per layer results in higher CERs compared to using 210 cells.
A network of this size is too small for modelling the acoustics entirely, but we are using it only as part of our superstructure to extract language dependent features.
On the other hand, having source nets with too many parameters may render the superstructure prone to over fitting.
Comparing the CERs of nets trained on graphemes and phonemes, the error rate of the German and Turkish grapheme based setup is lower in comparison to their phoneme based counterpart.
Potential reasons are a) the pronunciation dictionaries may be of varying quality as they were generated completely automatic and b) Turkish as well as German have easier pronunciation rules than English or French.
\begin{table}[h!]
\centering
\begin{tabular}{l|l|c|c|c|c}
\toprule
\textbf{Type} & \textbf{Size} & \textbf{DE} & \textbf{EN} & \textbf{FR} & \textbf{TR} \\
\midrule
Phone & 105    & 9.0 & 14.3 & 11.7 & 7.0 \\
Phone & 210   & 7.2 & 12.2 & 8.6 & 5.9 \\
\midrule
Grapheme & 105   & 8.3 & 16.5 & 13.3 & 7.2 \\
Grapheme & 210   & 6.4 & 13.1 & 9.6 & 5.6 \\
\bottomrule
\end{tabular}
\caption{CER of monolingual subnets}
\label{tab:subnets}
\end{table}
\subsection{Multilingual Systems}
We evaluated our multilingual setup using English data, with WERs shown in Table \ref{tab:lmdecode}.
As baseline experiment, we trained a system using English data only (setup 1).
Training the same setup jointly on data from multiple languages (setup 2) increases the WER on English.
Applying language codes via modulation (setup 3, \cite{mueller2018icassp}) improves the WER to 26.3\%.
Pre-training the networks using phonetic information from other but not the target language further reduces to WER to 25.4\% (setup 4, \cite{essv2018}), which almost achieves parity with the monolingual baseline (setup 1).
Switching then to our network superstructure based on Meta-PI and applying adaptive neural language codes (setup 5), the WER decreases further. 
This system not only achieves parity with the monolingual baseline, but even surpasses it.

On the best results, we then applied an optimized language model, which reduced the WERs of both systems (6 and 7), with the adapted system (7) again outperforming the monolingual setup (6).
  \begin{table}[h!]
    \centering
    \begin{tabular}{l|c}
      \toprule
      \textbf{Setup} & \textbf{WER} \\
      \midrule
      1) Monolingual baseline & 25.3\% \\
      \midrule
      2) No adaptation & 27.4\% \\
      3) LFV Modulation & 26.3\% \\
      4) Phonetic pre-training & 25.4\% \\
      \textbf{5) Meta-PI + NLC} & \textbf{24.2\%} \\
      \midrule \midrule
      6) Monolingual baseline (new LM) & 24.2\% \\
      \midrule
      7) \textbf{Meta-PI + NLC (new LM)} & \textbf{23.5\%} \\
      \bottomrule
    \end{tabular}
    \caption{WER of network superstructure, evaluated on English}
    \label{tab:lmdecode}
  \end{table}
\section{Conclusion}
\label{sec:conclusion}
We presented a language adaptation method for multilingual RNN/CTC based speech recognition systems.
It enables systems with a multilingual acoustic model to not only achieve parity with monolingual systems, but also to improve beyond the monolingual baseline.
Training acoustic models on data from multiple languages is a challenging problem because the acoustic modelling units between languages may differ.
This is especially problematic if graphemes are used as acoustic modelling units.
But by applying neural adaptation techniques proposed here, we are able to mitigate these issues and enable the adaptation of the acoustic model to multiple languages with not only achieving the same performance than the monolingual baseline, but also exceeding it.

Future work includes the integration of more languages into the setup, which we expect will allow for better generalization across languages.
But also the use of other acoustic units like, e.g., BPE (byte pair encoding) units should provide better results.
We also investigate the use of the proposed adaptation method to new domains. 
\bibliographystyle{IEEEtran}
\bibliography{mybib}
\end{document}